\documentclass{article}

\usepackage{microtype}
\usepackage{graphicx}
\usepackage{subfigure}
\usepackage{booktabs} 
\usepackage{amssymb}
\usepackage{amsmath}
\usepackage{extarrows}
\usepackage[normalem]{ulem}

\usepackage{hyperref}

\newcommand{\eg}{\emph{e.g.}}
\newcommand{\ie}{\emph{i.e.}}

\usepackage[accepted]{icml2019}

\icmltitlerunning{Addressing the Loss-Metric Mismatch with Adaptive Loss Alignment}

\begin{document}

\twocolumn[
\icmltitle{Addressing the Loss-Metric Mismatch with Adaptive Loss Alignment}




\begin{icmlauthorlist}
\icmlauthor{Chen Huang}{apple}
\icmlauthor{Shuangfei Zhai}{apple}
\icmlauthor{Walter Talbott}{apple}
\icmlauthor{Miguel Angel Bautista}{apple}
\icmlauthor{Shih-Yu Sun}{apple}
\icmlauthor{Carlos Guestrin}{apple}
\icmlauthor{Josh Susskind}{apple}
\end{icmlauthorlist}

\icmlaffiliation{apple}{Apple Inc., Cupertino, United States}

\icmlcorrespondingauthor{Chen Huang}{chen-huang@apple.com}

\icmlkeywords{ICML, Meta Learning, Reinforcement Learning}

\vskip 0.3in
]



\printAffiliationsAndNotice{}  

\begin{abstract}

In most machine learning training paradigms a fixed, often handcrafted, loss function is assumed to be a good proxy for an underlying evaluation metric. In this work we assess this assumption by meta-learning an adaptive loss function to directly optimize the evaluation metric. We propose a sample efficient reinforcement learning approach for adapting the loss dynamically during training. We empirically show how this formulation improves performance by simultaneously optimizing the evaluation metric and smoothing the loss landscape. We verify our method in metric learning and classification scenarios, showing considerable improvements over the state-of-the-art on a diverse set of tasks. Importantly, our method is applicable to a wide range of loss functions and evaluation metrics. Furthermore, the learned policies are transferable across tasks and data, demonstrating the versatility of the method. 

\end{abstract}

\section{Introduction}
\begin{figure}[!t]
\begin{center}
\centerline{\includegraphics[width=\columnwidth]{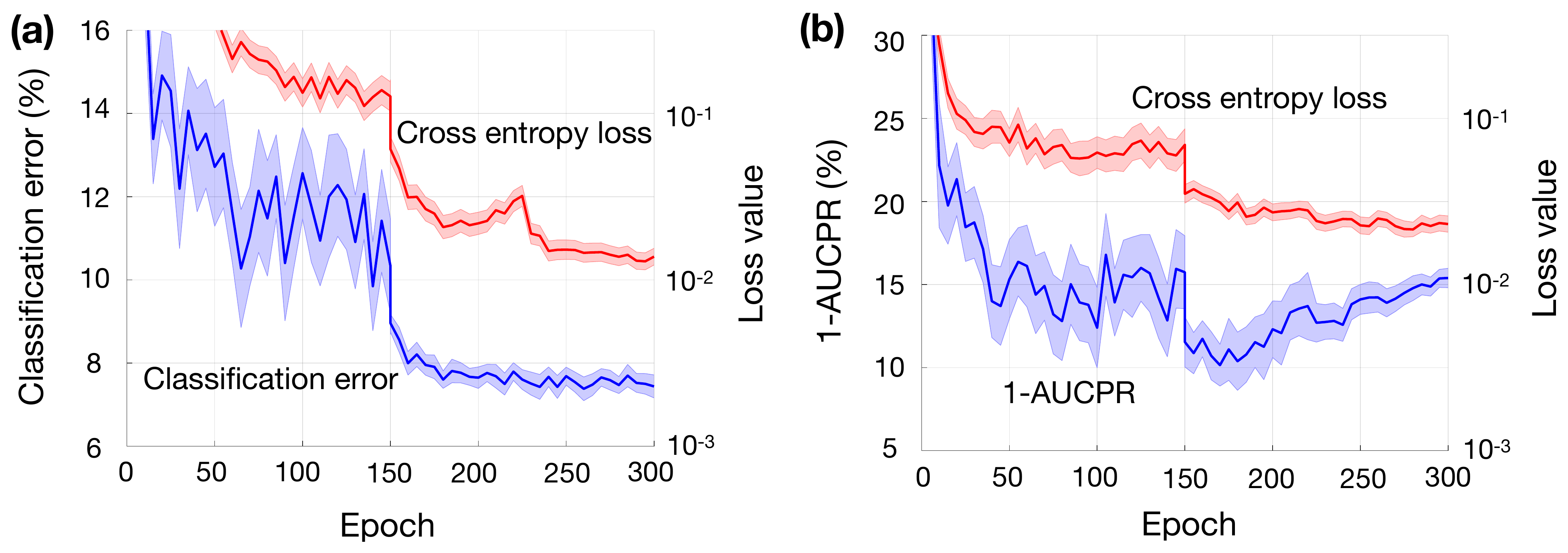}}
\vskip -0.1in
\caption{Loss-metric mismatch on CIFAR-10: (a) cross-entropy loss on the training set vs. classification error on the test set, (b) cross-entropy loss on the training set vs. area under the precision recall curve (AUCPR) on the test set. All curves are averaged over 10 runs initialized with different random seeds, with shaded areas showing the standard deviations. Note how training loss and evaluation metric curves differ in shape and noise characteristics, which is more apparent in the AUCPR case.}
\label{fig:Motivation}
\end{center}
\vskip -0.4in
\end{figure}

Machine learning models, including deep neural networks, are difficult to optimize, particularly for real world performance. One critical reason is that default loss functions are not always good approximations to evaluation metrics, a phenomenon we term the \emph{loss-metric mismatch} (see examples in Figure~\ref{fig:Motivation}). In fact, loss functions are often designed to be differentiable, and preferably convex and smooth, whereas many evaluation metrics are not. This problem is particularly evident in tasks like metric learning where the evaluation metrics of interest include area under the ROC curve (AUC) or precision/recall rate. These evaluation metrics are non-decomposable (involving statistics of a set of examples), nonlinear and non-continuous w.r.t. the sample-wise predictions and labels. One intuitive remedy is to experiment with many losses and identify one that correlates most to the evaluation metric, but this is inefficient both in terms of computation and human effort to design new losses.

The ideal choice of a loss function should tightly approximate the metric across a wide range of parameter values. During model training, the distribution of predictions tend to be different early in training compared to later when close to convergence. For instance, the softmax outputs of a classification model usually change gradually from a near uniform distribution to a sharp multinomial distribution. This poses yet another challenge to the ability of the default loss function to approximate the metric, and indicates the potential benefit of choosing an adaptive loss function, where its parameters are dynamically adjusted to serve as a better surrogate to the evaluation metric.


In addition to the loss-metric mismatch, another difficulty lies in the potential difference between the distribution of the training set and that of the test set, which can be due to different levels of data noise or sampling biases~\cite{RenZYU18}. Optimizing with respect to the training set thus can exhibit different characteristics from the test set, which will lead to a \emph{generalization gap}.

We address the loss-metric mismatch by introducing Adaptive Loss Alignment (ALA), a technique that automatically adjusts loss function parameters to directly optimize the evaluation metric on a validation set. We also find this results in an empirical reduction of the generalization gap. ALA learns to adjust the loss function using Reinforcement Learning (RL) at the same time as the model weights are being learned by gradient descent. This helps to align the loss function to the evaluation metric cumulatively over successive training iterations.


We experiment with a variety of classification and metric learning problems. Results demonstrate significant gains of our approach over fixed loss functions and other meta-learning methods. Furthermore, ALA is generic and applies to a wide range of loss formulations and evaluation metrics. The learned loss control policy can be viewed as a data-driven curriculum that enjoys good transferability across tasks and data. In summary, the contributions of this work are as follows:

\vspace{-10pt}
\begin{itemize} \setlength{\itemsep}{0pt}
\item We introduce ALA, a sample efficient RL approach to address the loss-metric mismatch directly by continuously adapting the loss.
\item We provide RL formulations for metric learning and classification, with state of the art results relative to fixed loss and other meta-learning baselines.
\item We show the versatility of ALA for a wide range of loss functions and evaluation metrics, and demonstrate its transferability across tasks and data.
\item We empirically show through ablation studies that ALA improves optimization as well as generalization.
\end{itemize}
\vspace{-10pt}

\section{Related work}


Improving model training through efficient experimentation is an active area of research. This includes classic hyperparameter optimization~\cite{BergstraB12,Snoek2012} and more recent gradient-based approaches~\cite{Maclaurin2015,pmlr-v70-franceschi17a}. These techniques overlap with architecture meta learning, such as activation functions~\cite{Ramachandran2017SearchingFA} and neural architecture search, \eg,~\cite{xie2018snas}. In addition, certain design choices can aid in generalization such as the use of batch normalization~\cite{IoffeS15, NIPS2018_7515}. ALA focuses on the mechanics of the loss function, and can be used independently or in conjunction with other methods to improve performance.

A different strategy is to focus on the optimization process itself and the associated dynamics of learning. Curriculum learning aims to improve optimization by gradually increasing the difficulty of training~\cite{Bengio2009}, which has been extended from the pre-defined case, \eg, focal loss~\cite{lin2017focal}, to learning a data-driven curriculum~\cite{JiangZLLF18} or a data reweighting scheme~\cite{RenZYU18}, by optimizing a proxy loss. Similarly, optimizers can be learned in a data driven fashion using gradient descent, \eg,~\cite{NIPS2016_6461, WichrowskaMHCDF17}. In contrast, ALA is a more general method since it adapts the loss function, which allows it to find better optimization strategies to address the loss-metric mismatch. 

Other methods focus on the loss function itself. These include approximate methods that bridge the gap between loss functions and evaluation metrics for special cases, \eg, for ranking losses/area under the curve~\cite{pmlr-v54-eban17a, NIPS2014_5367}. Closest to our work are learning to teach methods that adapt loss functions to optimize target metrics. In~\cite{xu2018autoloss}, reinforcement learning is used to learn a discrete optimization schedule that alternates between different loss functions at different time-points, which is suited to multi-objective problems, but does not address more general non-discrete cases. In~\cite{NIPS2018_7882} and~\cite{Jenni_2018_ECCV}, gradient-based techniques are used to optimize differentiable proxies for the evaluation metric, whereas our method optimizes the metric directly. 

ALA differs from the above techniques by optimizing the evaluation metric directly via a sample efficient RL policy that iteratively adjusts the loss function. Our work also extends beyond classification to the metric learning case. 

\section{Adaptive Loss Alignment (ALA)}
We start by formally defining our learning problem, where we would like to improve an evaluation metric $\mathcal{M}(f_w,D_{val})$ on validation set $D_{val}$, for a parametric model $f_w:\mathcal{X} \to \mathcal{Y}$.
The evaluation metric $\mathcal{M}$ between the ground-truth $y$ and model prediction $f_w(x)$ given input $x$, can be either decomposable over samples (\eg,~classification error) or non-decomposable like area under the precision recall curve (AUCPR) and Recall@k. 
We learn to optimize for the validation metric and expect it to be a good indicator of the model performance $\mathcal{M}(f_w,D_{test})$ on test set $D_{test}$.

Optimizing directly for evaluation metrics is a challenging task. This is because the model weights $w$ are actually obtained by optimizing a loss function $l$ on training set $D_{train}$,~\ie,~by solving $\min_w \sum_{(x,y)\in D_{train}} l(f_w(x),y)$. However, in many cases the loss $l$ is only a surrogate of the evaluation metric $\mathcal{M}$, which can be non-differentiable w.r.t. $w$. Moreover, the loss $l$ is optimized on the training set $D_{train}$ instead of $D_{val}$ or $D_{test}$.

To address the above \emph{loss-metric mismatch} issue, we propose to learn an adaptive loss function $l_{\Phi}(f_w(x),y)$ with loss parameters $\Phi \in \mathbb{R}^u$. The goal is to align the adaptive loss with evaluation metric $\mathcal{M}$, on a held-out dataset $D_{val}$. This leads to an alternate direction optimization problem --- (1) find metric-minimizing loss parameters $\Phi$ and (2) update the model weights $w_{\Phi}$ under the resultant loss $l_{\Phi}$ by,~\eg,~Stochastic Gradient Descent (SGD). We have:
\begin{equation}  
\begin{split}
\min_{\Phi} \;\;\; \mathcal{M}(f_{w_{\Phi}},D_{val}), \\
s.t. \; w_{\Phi} = \arg\min_w \sum_{(x,y)\in D_{train}} l_{\Phi}(f_{w}(x),y),
\end{split}
\label{eq1}
\end{equation}
where in practice both the outer loop and inner loop are approximated by a few steps of iterative optimization (\eg,~SGD updates). Hence, we denote $\Phi_t$ as the loss function parameters at time step $t$ and $w_{\Phi_t}$ as the corresponding model parameters. The key here is to bridge the gap between evaluation metric $\mathcal{M}$ and loss function $l_{\Phi_t}$ over time, conditioned on the found local optimum $w_{\Phi_t}$.

\begin{figure}[!t]
\vskip 0.2in
\begin{center}
\centerline{\includegraphics[width=\columnwidth]{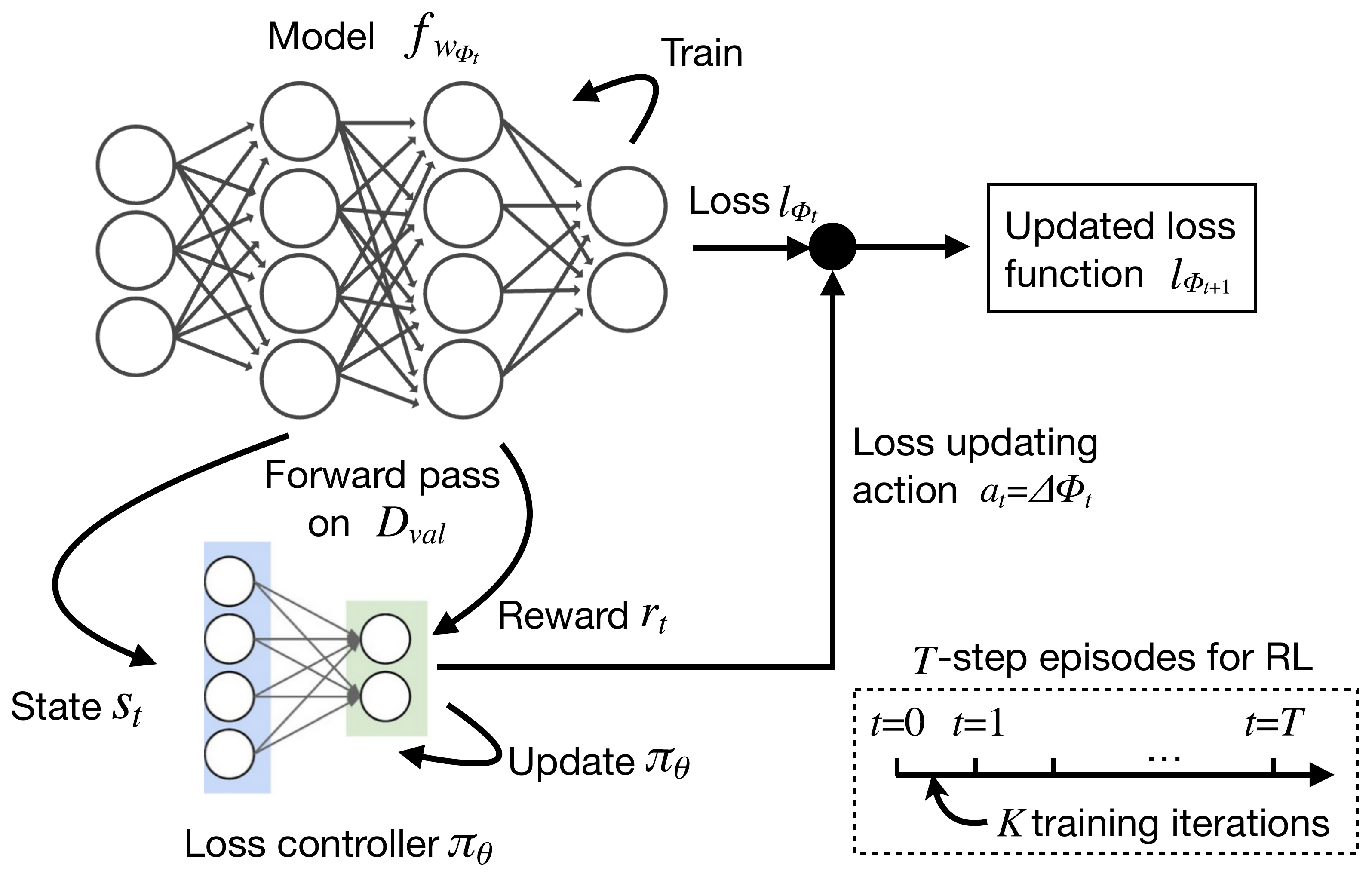}}
\caption{Reinforcement Learning (RL) of Adaptive Loss Alignment (ALA) controller. Model $f_{w_{\Phi_t}}$ is trained under loss function $l_{\Phi_t}$ with time-varying parameters $\Phi_t$. Given the model state $s_t$ at time $t$, the ALA controller proposes an action $a_t=\Delta \Phi_t$ to update the loss for follow-up training. The improvement in some evaluation metric (on validation set $D_{val}$) is used as reward $r_t$ to update the policy $\pi_\theta$. Our ALA controller is trained with one-step episodes (\ie,~$T=1$) and is sample efficient.}
\label{fig:Framework}
\end{center}
\vskip -0.4in
\end{figure}

\subsection{Reinforcement Learning of ALA}
To capture the conditional relations between loss and evaluation metric, we formulate a reinforcement learning problem. The task is to predict the best change $\Delta \Phi_t$ to loss parameters $\Phi_t$ such that optimizing the adjusted loss aligns better with the evaluation metric $\mathcal{M}(f_{w_{\Phi_t}},D_{val})$. In other words, taking an action that adjusts the loss function should produce a reward that reflects how much the metric $\mathcal{M}$ will improve on held-out data $D_{val}$. This is analogous to teaching the model how to better optimize on seen data $D_{train}$ and to better generalize (in terms of $\mathcal{M}$) on unseen data $D_{val}$.
 
The underlying model behind Reinforcement Learning (RL) is a Markov Decision Process (MDP) defined by states $s\in \mathcal{S}$ and actions $a\in \mathcal{A}$ at discrete time steps $t$ within an episode. In our case, an episode is naturally defined as a set of $T$ consecutive time-steps, each consisting of $K$ training iterations. In other words, the RL policy collects episodes at a slower timescale than the training of $f_w$. Figure~\ref{fig:Framework} illustrates the schematic of our RL framework. Our state $s_t$ records training progress information (\eg,~$w_{\Phi_t}$-induced loss value), and the loss-updating action $a_t=\Delta \Phi_t$ is sampled from a stochastic policy $\pi_\theta(a_t|s_t)$. We implement the policy as a neural network parameterized by $\theta$. Training $f_w$ under the updated loss $l_{\Phi_{t+1}}$ will transition to a new state $s_{t+1}$ and produce a reward signal $r_t = r(s_t, a_t)$.  We define the reward by the improvement in evaluation metric $\mathcal{M}(f_{w_{\Phi_t}},D_{val})$.

We optimize the loss-controlling policy with a policy gradient approach, similar to REINFORCE~\cite{Williams92}.  The objective is to maximize the expected total return 
\begin{equation}
J(\theta) = \mathbb{E}_{\tau} \left[ R(\tau) \right],
\label{eq2}
\end{equation}
where $R(\tau)$ is the total return $R(\tau) = \sum_{t=0}^{T} r_t$ of an episode $\tau = \{s_t, a_t | t \in [0, T]\}$. The updates to the policy parameters $\theta$ are given by the gradient
\begin{equation}
	\nabla_{\theta} J(\theta) = \mathbb{E}_{\tau} \left[ \sum_{t=0}^T \nabla_{\theta} \log \pi_\theta(a_t | s_t) \sum_{k=t}^T (r_k - b_k)\right],
\label{eq3}
\end{equation}
 where $b_k$ is a variance-reducing baseline implemented as the exponential moving average of previous rewards.

\textbf{Local policy learning.} As shown in Figure~\ref{fig:Framework}, an episode for RL is a set of $T$ consecutive time steps. Choosing one extreme, $T$ could cover the entire length of a training run. However, waiting for the model to fully train and repeating this process enough times for a policy to converge requires a great deal of computational resources. We choose the other extreme and use episodes of a single step,~\ie,~$T=1$ (still composed of $K$ training iterations). One advantage of doing so is that a single training run of $f_w$ can contribute many episodes to the training of $\pi_\theta$, making it sample efficient. Although this ignores longer-term effects of chosen actions, we will show empirically that these one-step episodes are sufficient to learn competent policies, and increasing $T$ does not convey much benefit in our experiments (see Supplementary Materials). The one-step RL setting is similar to a contextual bandit formulation, although actions affect future states. Thus, unlike bandit formulations, the ALA controller can learn state-transition dynamics.


\subsection{Learning Algorithm}
\label{sec:algo}



In this section, we will describe the concrete RL algorithm for simultaneous learning of the ALA policy parameters $\theta$ and model weights $w$.

\textbf{Rewards:} Our reward function $r_t$ measures the relative reduction in validation metric $\mathcal{M}(f_{w_{\Phi_t}},D_{val})$, after $K$ gradient descent iterations with an updated loss function $l_{\Phi_{t+1}}$. To represent the cumulative performance between model updates, we define:
\begin{equation}
\mathcal{M}_{t+1}=\sum_{j=1}^{K} \gamma^{K-j} \mathcal{M}(f_{w^{j}},D_{val}),
\label{eq4}
\end{equation}
where $\gamma$ is a discount factor that weighs more on the recent metric. The main model weights $w^{j}$ are updated for $K$ iterations. Then we quantize the reward $r_t$ to $\pm 1$ as follows:
\begin{equation}
r_t =\mathrm{sign}(\mathcal{M}_t-\mathcal{M}_{t+1}),
\label{eq5}
\end{equation}
which encourages continuous error metric decreases regardless of magnitude (but until the maximum training iteration).

\textbf{Action space:} For every element $\Phi_t(i)$ of the loss parameters $\Phi_t$, we sample action $a_t(i)$ from the discretized space $\mathcal{A}=\{-\beta,0,\beta\}$, with $\beta$ being a predefined step-size. Actions will update the loss parameters $\Phi_t(i) = \Phi_{t-1}(i) + a_t(i)$ at each time step. Our policy network $\pi_\theta$ has $|\mathcal{A}|$ output neurons for each loss parameter, and $a_t$ is sampled from a softmax distribution over these neurons. 

\textbf{State space:} Our policy network state $s_t \in \mathcal{S}$ consists of four components:
\begin{enumerate}

\item \vskip -0.1in 
Some task dependent validation statistics $S(f_{w_{\Phi_t}}, D_{val})$,~\eg,~the log probabilities of different classes, observed at multiple time-steps $\{t,t-1,\dots\}$.
\item The relative change $\Delta S(f_{w_{\Phi_t}}, D_{val})$ of validation statistics from their moving average.
\item The current loss parameters $\Phi_t$.
\item The current iteration number normalized by the total iteration number of the full training run of $f_w$.
\end{enumerate}
\vskip -0.1in

Here we use the validation statistics, among others, to capture model training states. Recall our goal is to find rewarding loss-updating actions $a_t=\Delta \Phi_t$ to improve the evaluation metric on validation set. A successful loss control policy would be able to model the implicit relation between the validation statistics, which is the state of the RL problem, and validation metric, which is the reward. In other words, ALA should learn to mimic the loss optimization process for decreasing the validation metric cumulatively. We choose to use validation statistics instead of training statistics in $s_t$ because the former is a natural proxy of the validation metric. Note the validation statistics are normalized in our state representation. This allows for generic policy learning which is independent of the actual model predictions from different tasks or loss formulations.

\begin{algorithm}[tb]
   \caption{Reinforcement Learning for ALA}
   \label{alg1}
\begin{algorithmic}
   \STATE Initialize each child model with random weights $w$
   \STATE Initialize loss controller $\pi_\theta$ with random weights $\theta$
   \STATE Initialize loss parameters $\Phi_0$ properly for a given task
   \STATE Initialize replay memory $\mathcal{D}$
   \WHILE{not converged}
   \FOR{each state $s_t$}
   \STATE Sample action $a_t \sim \pi_\theta(a_t|s_t)$
   \STATE Take action $a_t$ to update loss function to $l_{\Phi_{t+1}}$
   \STATE Update $w$ by $K$ SGD iterations with $l_{\Phi_{t+1}}$ 
   \STATE Collect reward $r_t$ (Equation~\ref{eq5}) and new state $s_{t+1}$
   \STATE Store $\langle s_t, a_t, r_t, s_{t+1} \rangle$ from all child models in $\mathcal{D}$
   \STATE Sample random experiences from $\mathcal{D}$
   \STATE Update $\theta$ to maximize reward via Equation~\ref{eq3}
   \ENDFOR
   \ENDWHILE
\end{algorithmic}
\label{alg1}
\end{algorithm}

\textbf{Algorithm:} Algorithm~\ref{alg1} shows how our RL algorithm alternates between updating the loss control policy $\pi_\theta$ and updating model weights $w$. Model weights are updated via mini-batch SGD on $D_{train}$, while policy $\pi_\theta$ is updated every $K$ SGD iterations on $D_{val}$. We enhance policy learning by training in parallel multiple main networks, which we refer to as child models. Each child model is initialized with random weights $w$ but sharing the loss controller. At each time-step $t$ for policy update, we collect the episodes using $\pi_\theta$ from all the child models. This independent set of episodes, together with replay memory $\mathcal{D}$ that adds randomness to the learning trajectories, help to alleviate the non-stationarity in online policy learning. As a result, more robust policies are learned, as verified in our experiments. 

It is worth noting that the initial loss parameters $\Phi_0$ are important for efficient policy learning. Proper initialization of $\Phi_0$ must ensure default loss function properties that depend on the particular form of loss parameterization for a given task,~\eg,~identity class correlation matrix in classification (see Section~\ref{sect:instantiation}).

\section{Instantiation in Typical Learning Problems}
\label{sect:instantiation}

\textbf{Classification:} We learn to adapt the parametric classification loss function introduced in~\cite{NIPS2018_7882}:
\begin{equation}
l_{\Phi_t}(f_w(x),y) = -\sigma(y^\mathrm{T} \Phi_t \log f_w(y|x)),
\label{eq6}
\end{equation}
where $\sigma(\cdot)$ is the sigmoid function, and $\Phi_t \in \mathbb{R}^{|\mathcal{Y}| \times |\mathcal{Y}|}$ denotes the loss function parameters with $|\mathcal{Y}|$ being the number of classes. $y \in \{0, 1\}^{|\mathcal{Y}|}$ denotes the one-hot representation of class labels, and $f_w(y|x) \in \mathbb{R}^{|\mathcal{Y}|}$ denotes the multinomial model output. This adaptive loss function is a generalization of the cross-entropy loss $l_{ce}(f_w(x),y) = -y^\mathrm{T} \log f_w(y|x)$ where $\Phi_t$ is fixed as the identity matrix.

The matrix $\Phi_t$ encodes time-varying class correlations. A positive value of $\Phi_t(i,j)$ encourages the model to increase the prediction probability of class $j$ given ground-truth class $i$. A negative value of $\Phi_t(i,j)$, on the other hand, penalizes the confusion between class $i$ and $j$. Thus when $\Phi_t$ changes as learning progresses, it is possible to implement a \emph{hierarchical curriculum} for classification, where similar classes are grouped as a super class earlier in training, and discriminated later as training goes further along. To learn the curriculum automatically, we initialize $\Phi_0$ as an identity matrix (reduced to the standard cross-entropy loss in this case), and update $\Phi_t$ over time by the ALA policy $\pi_\theta$.


To learn to update $\Phi_t$, we first construct a confusion matrix $C(f_{w_{\Phi_t}}, D_{val}) \in \mathbb{R}^{|\mathcal{Y}| \times |\mathcal{Y}|}$ of model prediction on validation set $D_{val}$, and define
\begin{equation}
C_{i,j} = \frac{\sum_{d=1}^{|D_{val}|}{-I(y_d, i)\log{f_{w_{\Phi_t}}^j(x_d)}}}{\sum_{d=1}^{|D_{val}|}{I(y_d, i)}},
\label{eq7}
\end{equation}
where $I(y_d, i)$ is an indicator function outputing 1 when $y_d$ equals class $i$ and 0 otherwise.

We take a parameter efficient approach to update each loss parameter $\Phi_t(i, j)$ based on the observed class confusions $[C_{i,j}, C_{j, i}]$. In other words, the ALA loss controller collects the validation statistics $S(f_{w_{\Phi_t}},D_{val})$ only for class pairs at each time step $t$, in order to construct state $s_t$ for updating the corresponding loss parameter $\Phi_t(i, j)$ (see Figure S2 in Supplementary Materials). Different class pairs share the same controller, and we update $\Phi_t(i, j)$ and $\Phi_t(j, i)$ to the same value (normalized between $[-1,1]$) to ensure class symmetry. This implementation is much more parameter efficient than learning to update the whole matrix $\Phi_t$ based on $C$. Furthermore, it does not depend on the number of classes for a given task, thus enabling us to transfer the learned policy to another classification task with an arbitrary number of classes (see Section \ref{sec:transfer}).

\textbf{Metric Learning:} A metric learning problem learns a distance metric to encode semantic similarity. Typically, the resulting distance metric cannot directly cater to different, and sometimes contradicting performance metrics of interest (\eg,~verification vs. identification rate, or precision vs. recall). This gap is more pronounced in the presence of common techniques like hard mining that only have indirect effects on final performance. Therefore, metric learning can serve as a strong testbed for learning methods that directly optimize evaluation metrics.

The standard triplet loss~\cite{SchroffKP15} for metric learning can be formulated as:
\begin{equation}
l_{tri}(f_w(x_{i,i^+,i^-})) = \max \left( 0,F(d^+)-F(d^-)+\eta \right),
\label{eq8}
\end{equation}
where $F(d)=d^2$ is the squared distance function over both $d^+$ (distance between anchor instance $f_w(x_i)$ and positive instance $f_w(x_{i^+})$) and $d^-$ (distance between $f_w(x_i)$ and negative instance $f_w(x_{i^-})$), while $\eta$ is a margin parameter.

As~\cite{wu2017sampling} pointed out, the shape of distance function matters. The concave shape of $-d^2$ for negative distance $d^-$ will lead to diminishing gradients when $d^-$ approaches zero. Here we propose to reshape the distance function adaptively with two types of loss parameterizations.

For the first parametric loss, called \textbf{Distance mixture}, we adopt 5 different-shaped distance functions for both $d^+$ and $d^-$, and learn an linear combination of them via $\Phi_t$:
\begin{equation}
l_{\Phi_t} = \sum_{i=1}^5 \Phi_t(i) F_i^+(d^+)+ \sum_{i=1}^{5} \Phi_t(i+5) F_i^-(d^-),
\label{eq9}
\end{equation}
where $F_i^+(\cdot)$ and $F_i^-(\cdot)$ correspond to the increasing and decreasing distance functions to penalize large $d^+$ and small $d^-$ respectively. In this case $\Phi_t = \{\Phi_t(i)\} \in [0,1]^{10}$, and is initialized as a binary vector such that $\Phi_0$ selects the default distance functions $d^2$ and $0.5d^{-1}$ for $d^+$ and $d^-$. For RL, the validation statistics in state $s_t$ are simply represented by the computed distance $F_i^+(\cdot)$ or $F_i^-(\cdot)$, and our ALA controller updates $\Phi_t(i)$ accordingly. Supplementary Materials specify the forms of $F_i^+(\cdot)$ and $F_i^-(\cdot)$ and show that final performance is not very sensitive to their design choices. It is more important to learn their dynamic weightings.

Similar to the focal loss~\cite{lin2017focal}, we also introduce a \textbf{Focal weighting}-based loss formulation as follows:
\begin{equation}
\begin{split}
l_{\Phi_t} & = \frac{1}{\Phi_t(1)}\log\left[1+\sum_{i+} \exp \left(\Phi_t(1) \cdot (d_{i^+}^+ -\alpha) \right) \right] \\
& + \frac{1}{\Phi_t(2)}\log\left[1+\sum_{i-} \exp \left(-\Phi_t(2) \cdot (d_{i^-}^- -\alpha) \right) \right],
\label{eq10}
\end{split}
\end{equation}

where $\Phi_t \in \mathbb{R}^2$, and $d_{i^+}^+$ and $d_{i^-}^-$ denote the distance between anchor instance and the positive $i^+$ and negative $i^-$ instances in the batch. We use these distances as validation statistics for ALA controller to update $\Phi_t$. While $\alpha$ here is the distance offset.


\section{Implementation Details}
The main model architecture and maximum number of training iterations are task-dependent. In the parallel training scenario for our experiments, 10 child models were trained sharing the same ALA controller. The controller is instantiated as an MLP consisting of 2 hidden layers each with 32 ReLU units. Our state $s_t$ includes a sequence of validation statistics observed from past 10 time-steps. Table S4 in Supplementary Materials quantifies these choices.


The ALA controller is learned using the REINFORCE policy gradient method (which worked better than Q-learning variants in early experiments). We use a learning rate of 0.001 for policy learning. Training episodes are collected from all child networks every $K=200$ gradient descent iterations. We set the discount factor $\gamma=0.9$ (Equation~\ref{eq4}), loss parameter updating step $\beta =0.1$ and distance offset $\alpha =1$ (Equation~\ref{eq10}), but found that performance is robust to variations of these hyperparameters.

\section{Results}

We evaluate the proposed ALA method on classification and metric learning tasks. Our ALA controller is learned by multi-model training for both tasks unless otherwise stated.


\subsection{Classification} \label{sec:classification}
We train and evaluate ALA using two different evaluation metrics to demonstrate generality: (1) classification error, and (2) area under the precision recall curve (AUCPR). We experiment on CIFAR-10~\cite{Krizhevsky09learningmultiple} with 50k images for training and 10k images for testing. For training a loss controller, we divide the training set randomly into a new training set of 40k images and a validation set of 10k images. We use Momentum-SGD for training. The compared methods below use the full 50k training images and their optimal hyper-parameters.

\textbf{Optimizing classification error:} Table~\ref{tb:cifar10} reports classification errors using the popular network architectures: ResNet-32~\cite{He2016DeepRL}, Wide-ResNet (WRN)~\cite{ZagoruykoK16} and DenseNet~\cite{huang2017densely}. The self-paced method~\cite{NIPS2010_3923} is a predefined curriculum learning scheme based on example hardness. L-Softmax~\cite{liu2016large} is a strong hand-designed loss function. The recent state-of-the-art L2T-DLF~\cite{NIPS2018_7882} method adapts a loss function under the same formulation of Equation~\ref{eq6}. However, the objective of L2T-DLF is to minimize an evaluation metric surrogate that is gradient-friendly. ALA outperforms L2T-DLF using single network training, and improves further with multi-network training (default). The competitive performance of our single network RL training validates its sample and time efficiency. Other ALA baselines train with either random loss parameters $\Phi_t$ (identity matrix perturbed by 10\% noise), or use the prediction confusion matrix (Equation~\ref{eq7}) as $\Phi_t$. These baselines do not perform well, while ALA can \emph{learn} meaningful loss parameters to align with evaluation metrics. The L2T method~\cite{fan2018learning} similarly uses RL to optimize evaluation metrics, but requires 50 episodes of full training. In contrast, our ALA controller is trained while the main model is being trained, thereby allowing for over $50\times$ faster learning with even stronger error reduction.

\textbf{Optimizing AUCPR:} Unlike classification error, AUCPR is a highly-structured and non-decomposable evaluation metric. To demonstrate our ability to optimize different metrics, we change the reward for ALA to AUCPR as defined in~\cite{pmlr-v54-eban17a}. Table~\ref{tb:cifar10_aucpr} shows our method achieves the AUCPR metric of 94.9\% (10-run average) on CIFAR-10, outperforming the SGD-based optimization method~\cite{pmlr-v54-eban17a}. Our advantages are also evident over methods that do not optimize the metric of interest,~\eg,~via the pairwise AUCROC surrogate~\cite{Rakotomamonjy2004OptimizingAU} and cross-entropy loss that is a proxy for classification accuracy.

\begin{table}[t]
\caption{Classification error (\%) on CIFAR-10 dataset. 10-run average and standard deviation are reported for ALA.}
\label{tb:cifar10}
\begin{center}
\begin{small}
\vskip -0.16in
\resizebox{\linewidth}{!}{
\begin{tabular}{lccc}
\toprule
Method & ResNet-32 & WRN & DenseNet \\
\midrule
cross-entropy & 7.51 & 3.80 & 3.54\\
Self-paced~\cite{NIPS2010_3923} & 7.47 & 3.84 & 3.50\\
L-Softmax~\cite{liu2016large} & 7.01 & 3.69 & 3.37 \\
L2T~\cite{fan2018learning} & 7.10 & - & - \\
L2T-DLF~\cite{NIPS2018_7882} & 6.95 & 3.42 & 3.08 \\ \midrule
ALA (random matrix $\Phi_t$) & 8.23$\pm$0.41 & 4.69$\pm$0.28 & 4.15$\pm$0.33 \\
ALA (confusion matrix $\Phi_t$) & 7.42$\pm$0.04 & 3.74$\pm$0.02 & 3.55$\pm$0.02 \\ \midrule
ALA (single-network) & 6.85$\pm$0.09 & 3.39$\pm$0.04 & 3.03$\pm$0.04 \\
ALA (multi-network) & \textbf{6.79$\pm$0.07} & \textbf{3.34$\pm$0.04} & \textbf{3.01$\pm$0.02}\\
\bottomrule
\end{tabular}
}
\end{small}
\end{center}
\vskip -0.2in
\end{table}

\begin{table}[t]
\caption{AUCPR (\%) on CIFAR-10 dataset. All methods use the same model architecture as adopted in~\cite{pmlr-v54-eban17a}. 10-run average and standard deviation are reported as ALA result.}
\label{tb:cifar10_aucpr}
\vskip 0.15in
\begin{center}
\begin{small}
\resizebox{0.95\linewidth}{!}{
\begin{tabular}{lc}
\toprule
Method & AUCPR \\ \midrule
cross-entropy loss for optimizing accuracy & 84.6 \\
Pairwise AUCROC loss~\cite{Rakotomamonjy2004OptimizingAU} & 94.2 \\
AUCPR loss~\cite{pmlr-v54-eban17a} & 94.2 \\
ALA & \textbf{94.9$\pm$0.14} \\
\bottomrule
\end{tabular}
}
\end{small}
\end{center}
\vskip -0.2in
\end{table}

\subsection{Metric Learning} \label{sec:metriclearning}
To validate ALA on metric learning tasks, we perform image retrieval experiments on the Stanford Online Products (SOP) dataset~\cite{SongXJS16}, and face recognition (FR) experiments on the LFW dataset~\cite{LFWTech}. The SOP dataset contains 120,053 images of 22,634 categories. The first 10,000 and 1,318 categories are used for training and validation, and the remaining are used for testing. We optimize for the evaluation metric of average Recall@k on SOP. For LFW, we train for the verification accuracy and test on the LFW verification benchmark containing 6,000 verification pairs. Note both the average Recall@k and verification accuracy are non-decomposable evaluation metrics.



Table~\ref{tb:ml} (top two cells) compares ALA to recent methods on the SOP dataset. ALA is applied to two representative metric learning frameworks---Triplet~\cite{SchroffKP15} and Margin~\cite{wu2017sampling}, using the respective network architectures. The two works differ in the loss formulation and data sampling strategies. We can see that ALA leads to consistently significant gains at all recall levels for both frameworks. Margin+ALA outperforms recent embedding ensembles based on boosting (BIER) and attention (ABE-8) mechanisms, as well as HTL using a hand-designed hierarchical class-tree. This indicates the advantages of our adaptive loss function and direct metric optimization. On LFW, ALA improves the triplet framework with distance mixtures, and achieves state-of-the-art verification accuracy 99.57\% under the small training data protocol. Please refer to the Supplementary Materials for comparisons to recent strong FR methods on LFW.

\begin{table}[t]
\caption{Recall(\%)@k on Stanford Online Products dataset.}
\label{tb:ml}
\begin{center}
\begin{small}
\vskip -0.1in
\resizebox{\linewidth}{!}{
\begin{tabular}{lcccc}
\toprule
k & 1 & 10 & 100 & 1000 \\
\midrule
Triplet~\cite{SchroffKP15} & 66.7 & 82.4 & 91.9 & - \\
Margin~\cite{wu2017sampling} & 72.7 & 86.2 & 93.8 & 98.0 \\
BIER~\cite{Opitz17} & 72.7 & 86.5 & 94.0 & 98.0 \\
HTL~\cite{Ge18} &74.8 & 88.3 & 94.8 & 98.4 \\
ABE-8~\cite{Wonsik18} & 76.3 & 88.4 & 94.8 & 98.2 \\
\midrule
Triplet + ALA (Distance mixture) & 75.7 & 89.4 & 95.3 & 98.6 \\
Margin + ALA (Distance mixture) & \textbf{78.9} & \textbf{90.7} & \textbf{96.5} & \textbf{98.9} \\ \midrule
Margin + ALA (Focal weighting) & 77.9 & 90.1 & 95.8 & 98.7 \\ \midrule
Margin + ALA (FR policy transfer) & 75.2 & 89.2 & 94.9 & 98.4 \\ 

\bottomrule
\end{tabular}
}
\end{small}
\end{center}
\vskip -0.1in
\end{table}

\begin{table}[t]
\vskip -0.1in
\caption{Policy transfer from CIFAR-10 to ImageNet. Top-1 and Top-5 accuracy rates (\%) are reported on ImageNet.}
\label{tb:transfer_cls}
\begin{center}
\begin{small}
\resizebox{0.85\linewidth}{!}{
\begin{tabular}{lcc}
\toprule
Method & Top-1 & Top-5 \\
\midrule
RMSProp & 73.5 & 91.5 \\
PowerSign-cd~\cite{pmlr-v70-bello17a} & 73.9 & 91.9 
\\\midrule
RMSProp + ALA (CIFAR policy transfer) & 74.3 & 92.1 \\
RMSProp + ALA & \textbf{74.6} & \textbf{92.6} \\
\bottomrule
\end{tabular}
}
\end{small}
\end{center}
\vskip -0.2in
\end{table}

\subsection{Transfer Learning} \label{sec:transfer}
One potential disadvantage of hand-designed loss functions compared to learned ones is that the former may be specific to task and/or data, whereas the latter are usually generic and widely applicable. Here we test the degree to which ALA extracts general knowledge about how to adapt the loss. We do so by conducting policy transfer experiments on classification and metric learning tasks in the following.


For classification, we transfer the learned policy that updates pairwise class correlations based on their prediction confusions. Specifically, we train an ImageNet~\cite{DenDon09Imagenet} classifier with RMSProp optimizer, but using the fixed ALA loss policy learned from CIFAR-10 (with DenseNet). Despite the difference between number of classes and input distribution, the policy transfer is straightforward thanks to the weight sharing design of the policy network, We compare with PowerSign-cd~\cite{pmlr-v70-bello17a} that transfers a learned policy \emph{for optimization} from CIFAR-10 to ImageNet. Table~\ref{tb:transfer_cls} compares all methods using identical NASNet-A network architectures. It shows that the transferred ALA policy outperforms the baselines without careful tuning, and is comparable to the ImageNet-tuned RMSProp+ALA. These results show that ALA is able to learn a policy on small, quick-to-train datasets in order to guide large-scale training, which is desirable for fast and efficient learning. Notably, we use a different optimizer---SGD---for training the ALA policy on CIFAR-10, which shows that ALA is robust to the optimizer paired with the loss policy.


We also show that policy transferability holds in the metric learning case. Here we transfer the ALA policy learned from FR experiments to guide the training on SOP for image retrieval. The bottom row of Table~\ref{tb:ml} shows that the transferred policy is competitive with specialized learning on the target data domain. This demonstrates the superior generalization ability of ALA.

\begin{table}[t]
\vskip -0.1in
\caption{Training and testing error (\%) of ResNet-32 on CIFAR-10. We report 10-run average and standard deviation results for ALA.}
\label{tb:ablation_cls}
\begin{center}
\begin{small}
\resizebox{\linewidth}{!}{
\begin{tabular}{lcc}
\toprule
Method & Training & Testing \\
\midrule

cross-entropy & 0.32 & 7.51 \\
ALA & \textbf{0.12$\pm$0.02} & \textbf{6.79$\pm$0.07} \\ \midrule
ALA-2nd run (using policy from 1st run) & 0.14$\pm$0.01 & 6.72$\pm$0.04 \\
ALA-2nd run (policy finetuning) & \textbf{0.08$\pm$0.01} & \textbf{6.72$\pm$0.02} \\
\bottomrule
\end{tabular}
}
\end{small}
\end{center}
\vskip -0.24in
\end{table}

\subsection{Analysis} \label{sec:analysis}

\textbf{Ablation studies:} \textbf{(i) Training vs. testing metric:} Table~\ref{tb:ablation_cls} (top cell) compares ALA with fixed cross-entropy loss in both training and testing classification errors. Results suggest that through dynamic loss-metric alignment, ALA improves both optimization on the training data and generalization on test data. \textbf{(ii) Time efficiency:} Table~\ref{tb:ablation_cls} (bottom cell) compares our default online policy learning against continuing to train with ALA for a second run (thus halving the learning efficiency). The compared baselines either reuse the learned policy in the 1st run or finetune the policy during the 2nd run. These baselines lead to marginal gains due to more learning episodes. On the other hand, online ALA suffices to learn competent policies. \textbf{(iii) Robustness to different loss parameterizations:} Table~\ref{tb:ml} (penultimate row) shows that ALA works similarly well when using a focal weighted loss function for the metric learning task. We conjecture that ALA can work with minimal tuning across a variety of parameterized loss formulations.


\begin{figure*}[!t]
\begin{center}
\centerline{\includegraphics[width=0.85\linewidth]{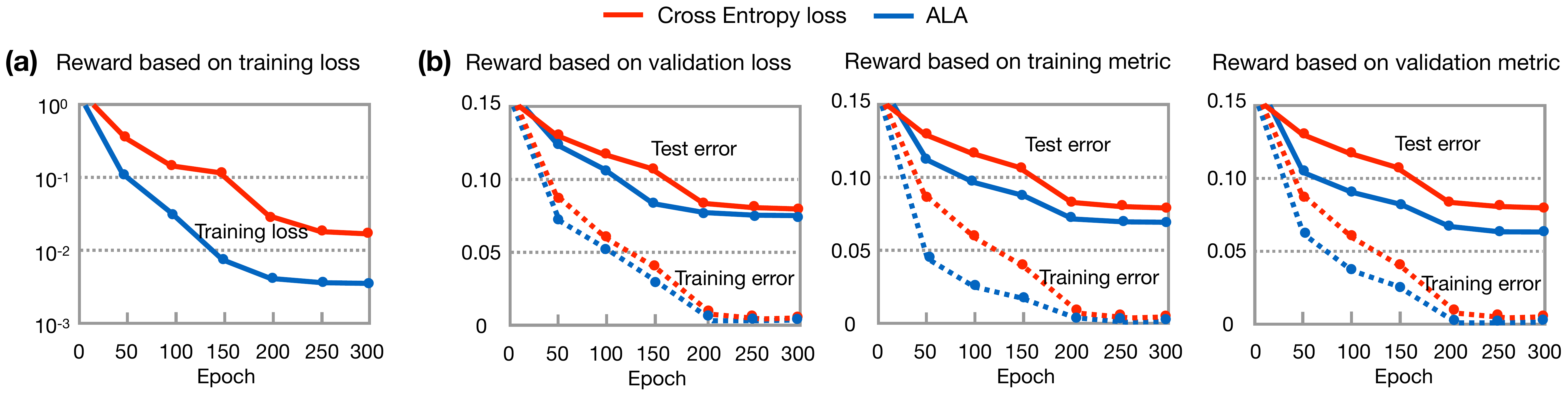}}
\vskip -0.1in
\caption{
Analysis of optimization vs. generalization on CIFAR-10. (a) Comparing optimization performance in terms of the raw cross-entropy loss outputs on training data: Here ALA is rewarded by the training loss, and we observe that the measured training loss is consistently lower compared to the fixed cross-entropy loss, indicating improved optimization. (b) Comparing ALA policies trained with different rewards: the validation loss-based reward improves both optimization (training error) and generalization (test error), and the gains are larger when using the validation metric-based reward. In contrast, using the training metric as the reward yields smaller gains in test error, potentially due to the diminishing reward (error approaching zero) in training data. 
}
\label{fig:Different_reward}
\end{center}
\vskip -0.3in
\end{figure*}

\textbf{Insights on optimization vs. generalization:} Figure~\ref{fig:Different_reward} analyzes the effects of ALA policies on optimization and generalization behavior by switching the reward signal in a 2-factor design: training vs. validation data and loss vs. evaluation metric on CIFAR-10. When the training loss (\ie, the default cross-entropy loss) or metric is used as reward, we compute them on the whole training set $D_{train}$. For this and the following studies, we train with the ResNet-32 architecture. 

Figure~\ref{fig:Different_reward}(a) isolates the optimization effects by comparing ALA, with the cross-entropy loss as the reward, and minimizing cross-entropy loss directly. ALA is shown to encourage better optimization with consistently lower loss values. We note that this is a remarkable result as it shows that ALA is facilitating optimization even when the objective is fully differentiable. Figure~\ref{fig:Different_reward}(b) examines both optimization and generalization when ALA uses validation loss as a reward where we monitor training and test (generalization) errors for ALA and fixed cross-entropy loss. We observe that the generalization error on test set is indeed improved, and the optimization in training error also sees small improvements. The optimization and generalization gains are larger when the validation metric is used as reward, which further addresses the loss-metric mismatch. By comparison, the training metric-based reward yields faster training error decrease, but smaller gains in test error potentially due to the diminishing error/reward in training data.


\begin{figure}[!t]
\begin{center}
\centerline{\includegraphics[width=0.7\columnwidth]{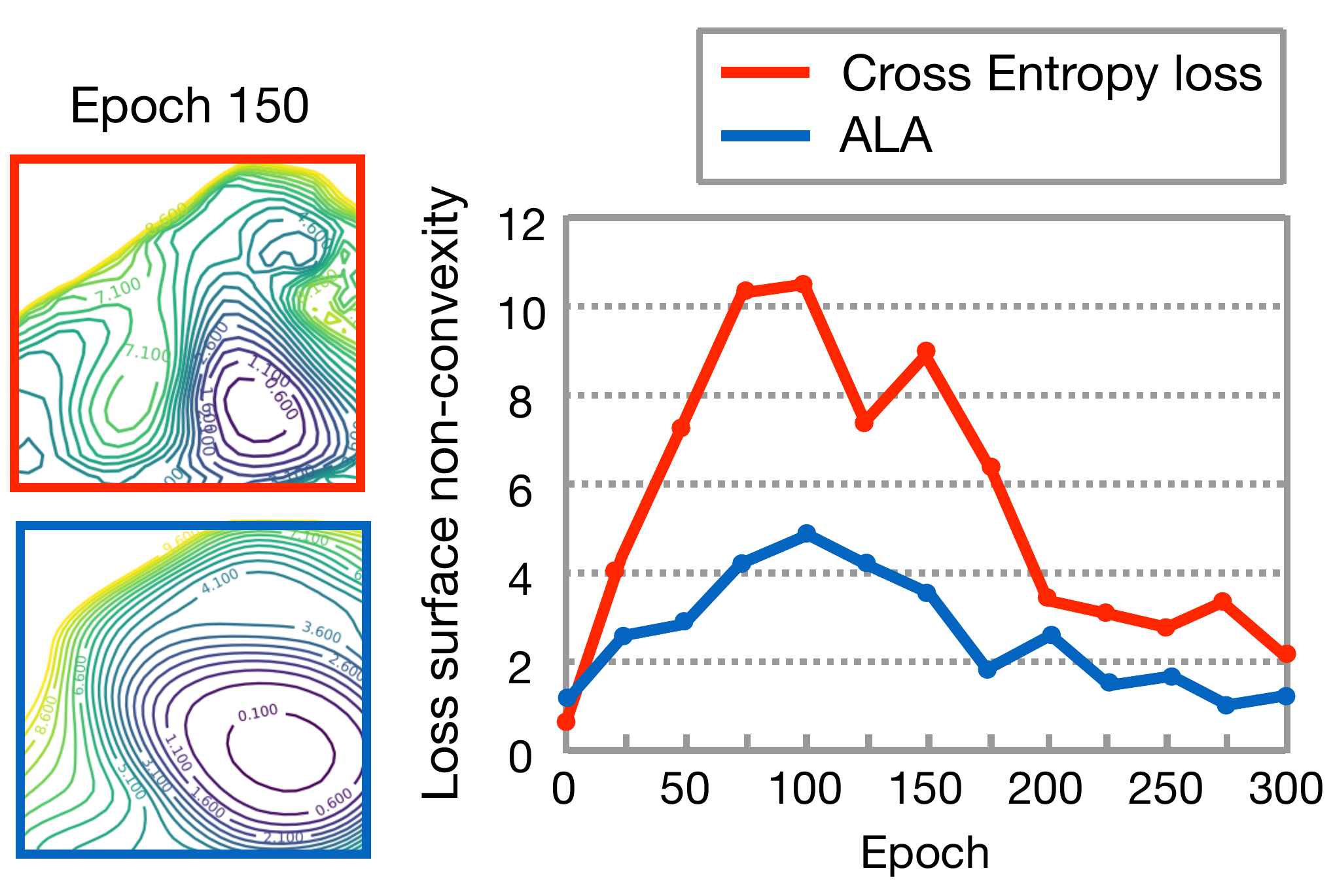}}
\caption{Analysis of the loss surface non-convexity on CIFAR-10: for both the fixed cross entropy loss and ALA, we compute the loss surface along filter-normalized 2D directions as in~\cite{visualloss} at different training epochs. Non-convexity is measured as the average Gaussian curvature for all points in the 2D loss surface.}
\label{fig:Loss_convexity}
\end{center}
\vskip -0.4in
\end{figure}

\textbf{Reasoning behind the improved optimization by ALA:} First, we speculate that ALA improves optimization by dynamically smoothing the loss landscape. We verify this hypothesis by taking measurements of the loss surface convexity, calculating the Gaussian curvature of the loss surface around each model checkpoint following~\cite{visualloss}. Figure~\ref{fig:Loss_convexity} shows a smoother loss surface from the model trained with ALA. This confirms that ALA is learning to manipulate the loss surface in a way that improves convergence of SGD based optimization processes, in agreement with findings in~\cite{visualloss}.

\begin{figure}[!t]
\begin{center}
\centerline{\includegraphics[width=1.05\columnwidth]{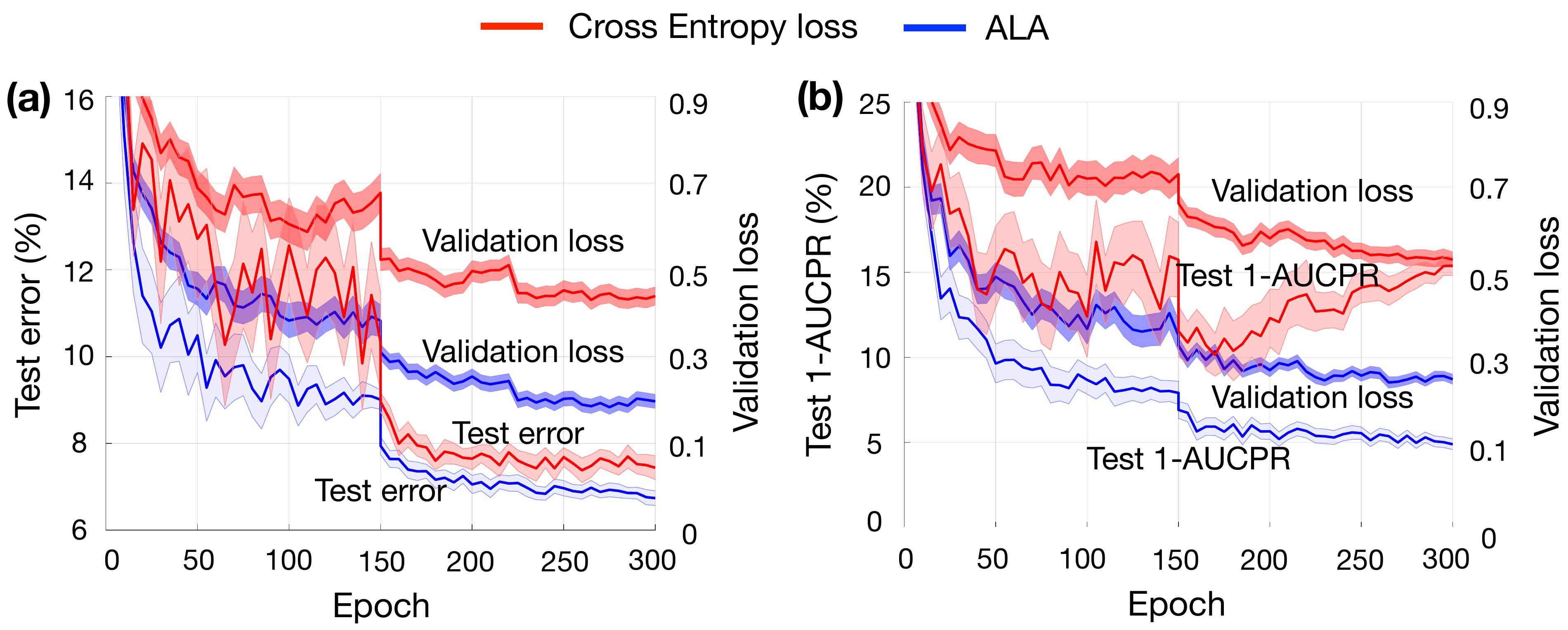}}
\caption{Validation loss vs. test metric of (a) classification error and (b) AUCPR on CIFAR-10. Curves are means over 10 runs initialized with different random seeds, and shaded areas show standard deviations. ALA uses the default reward based on the validation error metric, and improves both validation loss and test metric by addressing the loss-metric mismatch (both before and after the loss/metric drop due to learning rate change).}
\label{fig:Mismatch_visual}
\end{center}
\vskip -0.4in
\end{figure}

Second, we study how ALA improves performance by addressing the loss-metric mismatch. Figure~\ref{fig:Mismatch_visual} shows CIFAR-10 training with ALA using (a) classification error and (b) AUCPR metrics. We use the validation metric-based reward for ALA by default, and follow the aforementioned training settings for each metric. We can see that the fixed cross-entropy loss, without explicit loss-metric alignment, suffers from an undesirable mismatch between the monitored validation loss and test metric in both cases (see the different curve shapes and variances). ALA reduces the loss-metric mismatch by sequentially aligning the loss to the evaluation metric, even though classification error and AUCPR cases exhibit different patterns. In doing so, ALA not only lowers the validation loss but also the generalization test error.

\section{Conclusion}

We introduced ALA to make it easier to improve model performance on task-specific evaluation metrics. Performance is often hindered by the loss-metric mismatch, and we showed that ALA overcomes this by sequentially aligning the loss to the metric. We demonstrated significant gains over existing methods on classification and metric learning, using an efficient method that is simple to tailor, which makes ALA useful for automated machine learning. Intriguingly, ALA improves optimization and generalization simultaneously, in contrast to methods that focus on one or the other. We leave theoretical understanding of the effectiveness of loss function adaptation to future work.

\section*{Acknowledgements}
The authors want to thank (in alphabetical order) Leon Gatys, Kelsey Ho, Qi Shan, Feng Tang, Karla Vega, Russ Webb and many others at Apple for helpful discussions during the course of this project. In addition, we are grateful to Harry Guo, Myra Haggerty, Jerremy Holland and John Giannandrea for supporting the research effort. We also thank the ICML reviewers for providing useful feedback.

\bibliography{dynamic_loss}
\bibliographystyle{icml2019}

\newpage
\appendix

\setcounter{figure}{0}
\setcounter{table}{0}

\makeatletter 
\renewcommand{\thefigure}{S\@arabic\c@figure}
\renewcommand{\thetable}{S\arabic{table}}
\makeatother

\section*{Supplementary Material}

\section{More Experiments}

\textbf{Image classification on CIFAR-100:} We train and evaluate ALA for the metric of classification error on the CIFAR-100 dataset~\cite{Krizhevsky09learningmultiple}. As in CIFAR-10, we divide the training set of CIFAR-100 randomly into a new training set of 40k images and a validation set of 10k images, for loss controller learning. The 10k testing images are used for evaluation. We compare with the recent methods that use the full 50k training images and their optimal hyper-parameters. For ALA, multi-network training is adopted by default for robust online policy learning. Each network is trained via Momentum-SGD.

Table~\ref{tb:cifar100} reports classification errors using different ResNet~\cite{He2016DeepRL} architectures. For all network architectures, ALA outperforms both hand-designed loss functions,~\eg,~L-Softmax~\cite{liu2016large}, and the adaptive loss function that acts as a differentiable metric surrogate in L2T-DLF~\cite{NIPS2018_7882}. This validates the benefits of directly optimizing the evaluation metric using ALA.

\textbf{Face verification on LFW:} We evaluate the performance of our ALA-based metric learning method on a face verification task using the LFW dataset~\cite{LFWTech}. The LFW verification benchmark contains 6,000 verification pairs. For a fair comparison with recent approaches we train ALA using the same 64-layer ResNet architecture proposed in~\cite{liu2017sphereface,2018Wang} as our main model. We follow the small training data protocol~\cite{LFWTech} and train and validate on the popular CASIA-WebFace dataset~\cite{YiLLL14a} which contains 494,414 images of 10,575 people. The training images with identities appearing in the test set are removed. Our ALA controller is trained to optimize the verification accuracy metric on the validation set.

Table~\ref{tb:lfw} compares ALA to recent face recognition methods on LFW. These methods often adopt a strong but hand-designed loss function to improve class discrimination. In contrast, ALA adaptively controls the triplet loss function~\cite{SchroffKP15}, achieving state-of-the-art performance even for different parameterizations, where we specifically studied focal weighting~\cite{lin2017focal} and distance mixture formulations. These results further verify the advantages of ALA to directly optimize for the target metric regardless of the specific formulation of loss function to be controlled.

\begin{table}[t]
\caption{Classification error (\%) on CIFAR-100 dataset. 10-run average and standard deviation are reported for ALA.}
\label{tb:cifar100}
\begin{center}
\begin{small}
\vskip -0.16in

\resizebox{\linewidth}{!}{
\begin{tabular}{lccc}
\toprule
Method & ResNet-8 & ResNet-20 & ResNet-32 \\
\midrule
cross-entropy & 39.79 & 32.33 & 30.38\\
L-Softmax~\cite{liu2016large} & 38.93 & 31.65 & 29.56 \\
L2T-DLF~\cite{NIPS2018_7882} & 38.27 & 30.97 & 29.25 \\
ALA & \textbf{37.78$\pm$0.09} & \textbf{30.54$\pm$0.07} & \textbf{29.06$\pm$0.09}\\
\bottomrule
\end{tabular}
}
\end{small}
\end{center}
\vskip -0.1in
\end{table}

\begin{table}[t]
\caption{Face verification accuracy (\%) on LFW dataset. All methods use the same training data and network architectures.}
\label{tb:lfw}
\vskip 0.15in
\begin{center}
\begin{small}
\resizebox{0.85\linewidth}{!}{
\begin{tabular}{lc}
\toprule
Method & Accuracy \\ \midrule
Softmax loss & 97.88 \\
Softmax+Contrastive~\cite{Yi14} & 98.78 \\
Triplet loss~\cite{SchroffKP15} & 98.70 \\
L-Softmax loss~\cite{liu2016large} & 99.10 \\
Softmax+Center loss~\cite{wen2016} & 99.05 \\
SphereFace (A-Softmax)~\cite{liu2017sphereface} & 99.42 \\
CosFace (LMCL)~\cite{2018Wang} & 99.33 \\\midrule
Triplet + ALA (Focal weighting) & 99.49 \\
Triplet + ALA (Distance mixture) & \textbf{99.57} \\
\bottomrule
\end{tabular}
}
\end{small}
\end{center}
\vskip -0.1in
\end{table}

\begin{table}[!t]
\caption{Baseline comparisons for CIFAR-10 classification and metric learning on Stanford Online Products (SOP) dataset. We report classification error (\%) with ResNet-32 and Recall(\%)@k=1 for the two tasks respectively. For metric learning, ALA is trained with the `Margin' framework and with the loss parameterization of `Distance mixture'. We compare ALA to contextual bandit and population-based training (PBT) baselines.}
\label{tb:baseline}
\begin{center}
\begin{small}
\vskip -0.1in
\resizebox{\linewidth}{!}{
\begin{tabular}{lclc}
\toprule
\multicolumn{2}{c}{Classification} & \multicolumn{2}{c}{Metric learning} \\
\cmidrule(lr){1-2}
\cmidrule(lr){3-4}
Method & Error$\downarrow$ & Method & Recall$\uparrow$ \\ \midrule
cross-entropy & 7.51 & Triplet~\cite{SchroffKP15} & 66.7 \\
L2T~\cite{fan2018learning} & 7.10 & Margin~\cite{wu2017sampling} & 72.7 \\
L2T-DLF~\cite{NIPS2018_7882} & 6.95 & ABE-8~\cite{Wonsik18} & 76.3 \\
\midrule
ALA & \textbf{6.79} & Margin + ALA & \textbf{78.9} \\ \midrule
Contextual bandit & 7.34 & Contextual bandit & 73.1 \\
PBT~\cite{pbt2017} & 7.29 & PBT~\cite{pbt2017} & 73.6 \\

\bottomrule
\end{tabular}
}
\end{small}
\end{center}
\vskip -0.15in

\end{table}

\section{More Analyses}

\textbf{Baseline comparisons:} Table~\ref{tb:baseline} compares some related baselines in both classification and metric learning tasks to further highlight the benefits of ALA. In particular, we compare with the contextual bandit method and population-based training (PBT)~\cite{pbt2017}. The two baselines follow the same experimental settings on respective datasets, as detailed in the main paper.

\begin{figure}[!t]
\begin{center}
\centerline{\includegraphics[width=\columnwidth]{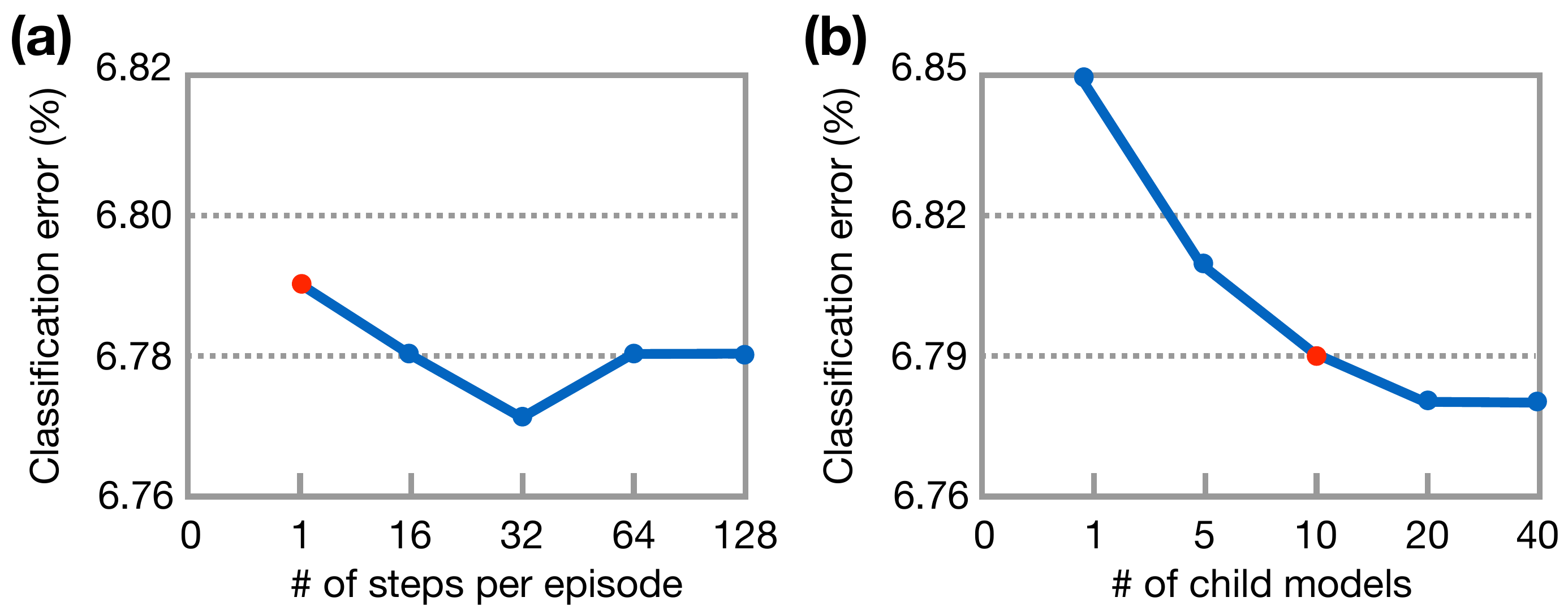}}
\caption{Sample efficiency of our RL approach for ALA (validation metric as reward). Classification error of ResNet-32 is reported on CIFAR-10. Good performance can be achieved by our default RL settings (red dots) with one-step episodes and 10 child model training that are sample efficient.}
\label{fig:Sample_efficiency_study}
\end{center}
\vskip -0.3in
\end{figure}

The contextual bandit method changes loss parameters (\ie,~actions) according to the current training states, similar to an online hyperparameter search scheme. Following the same loss parameterizations for classification and metric learning, the method increases weights for those confusing class pairs and evaluation metric-improving distance functions respectively, and otherwise downweights them. This is similar to our one-step RL setting except that in ALA, actions affect future states, making it an RL problem. Table~\ref{tb:baseline} illustrates that one-step RL-based ALA consistently outperforms the heuristic contextual bandit method. We believe more advanced bandit algorithms can work better, but RL has the capacity to learn flexible state-transition dynamics. Moreover, our RL setting can be extended to use multi-step episodes (Figure~\ref{fig:Sample_efficiency_study}). This allows to model longer-term effects of actions, while contextual bandits always obtain immediate reward from a single action.

Recall that we train in parallel 10 child models by default, for robust ALA policy learning. We are thus interested to see how this compares to PBT techniques (using the same 10 child models). Table~\ref{tb:baseline} shows that PBT does not help as much as ALA, which suggests the learned ALA policy is more powerful than model ensembling or parameter tuning. We will show later (in Figure~\ref{fig:Sample_efficiency_study}) that ALA can achieve competitive performance even with 1 child model which enjoys higher learning efficiency.

\textbf{Ablation study:} Table~\ref{tb:hyperparameter} shows the results of ablation studies on the design choices of \emph{ALA loss controller} and \emph{state representation}. As in Table~\ref{tb:baseline}, we experiment with the example tasks of classification and metric learning under the same settings. Looking at the top cell of Table~\ref{tb:hyperparameter}, we find that switching from 2-layer loss controller to 1-layer leads to a consistent performance drop; on the other hand, the 3-layer loss controller does not help much. The bottom cell of Table~\ref{tb:hyperparameter} quantifies the effects of the four components of our policy state $s_t$. We can see that it is relatively more important to keep the historical sequence of validation statistics (besides the ones at current timestep) and the current loss parameters $\Phi_t$ in the state representation. The relative change of validation statistics (from their moving average) and the normalized iteration number also have marginal contributions.

\textbf{Computational cost:} Under the classification and metric learning tasks considered in the paper, our simultaneous (single) model training and ALA policy learning often incur an extra $20\%-50\%$ cost as indexed by wall-clock time over regular model training. However, this overhead is often canceled out by the convergence speedup of the main model. Our multi-model training together with policy learning is able to achieve stronger performance with modest additional ($\sim30\%$) computational overhead for policy learning, at the cost of using distributed training to collect replay episodes. This is much more efficient than those meta-learning methods,~\eg,~\cite{fan2018learning,Barret2017} that learn the policy by training the main model to convergence multiple times (\eg,~50 times).

\begin{table}[t]
\caption{Ablation studies of ALA loss controller design (2-layer MLP by default) and state representation $s_t$. Experiments of CIFAR-10 classification and metric learning on SOP are conducted with the same settings as in Table~\ref{tb:baseline}. Performance degradation in comparison to default ALA method is indicated by positive $\Delta$ of classification error (\%) and negative $\Delta$ of Recall(\%)@k=1.}
\label{tb:hyperparameter}
\begin{center}
\begin{small}
\vskip -0.1in
\resizebox{\linewidth}{!}{
\begin{tabular}{lclc}
\toprule
\multicolumn{2}{c}{Classification} & \multicolumn{2}{c}{Metric learning} \\
\cmidrule(lr){1-2}
\cmidrule(lr){3-4}
Method & $\Delta$Error & Method & $\Delta$Recall \\ \midrule
ALA (1-layer MLP) & +0.06 & Margin+ALA (1-layer MLP) & -0.5 \\
ALA (3-layer MLP) & -0.03 & Margin+ALA (3-layer MLP) & -0.1 \\ 
\midrule
ALA ($s_t$ w/o history) & +0.11 & Margin+ALA ($s_t$ w/o history)  & -1.4 \\
ALA ($s_t$ w/o $\Delta$statistics) & +0.04 & Margin+ALA ($s_t$ w/o $\Delta$statistics) & -0.2 \\ 
ALA ($s_t$ w/o $\Phi_t$) & +0.05 & Margin+ALA ($s_t$ w/o $\Phi_t$) & -0.6 \\ 
ALA ($s_t$ w/o iter\#) & +0.02 & Margin+ALA ($s_t$ w/o iter\#) & -0.3 \\ 
\bottomrule
\end{tabular}
}
\end{small}
\end{center}
\end{table}

\textbf{Sample efficiency:} Figure~\ref{fig:Sample_efficiency_study} illustrates the sample efficiency of ALA's RL approach in the example task of CIFAR-10 classification. We train the ResNet-32 model and use the default reward based on the validation metric. Figure~\ref{fig:Sample_efficiency_study}(a) shows that using episodes consisting of a single training step suffices to learn competent loss policies with good performance. Figure~\ref{fig:Sample_efficiency_study}(b) further shows improvements from parallel training with multiple child models that provide more episodes for policy learning. We empirically choose to use 10 child models, which only incurs an extra $\sim30\%$ time cost for policy learning, thus striking a good performance tradeoff.

\begin{figure}[!t]
\begin{center}
\centerline{\includegraphics[width=0.55\columnwidth]{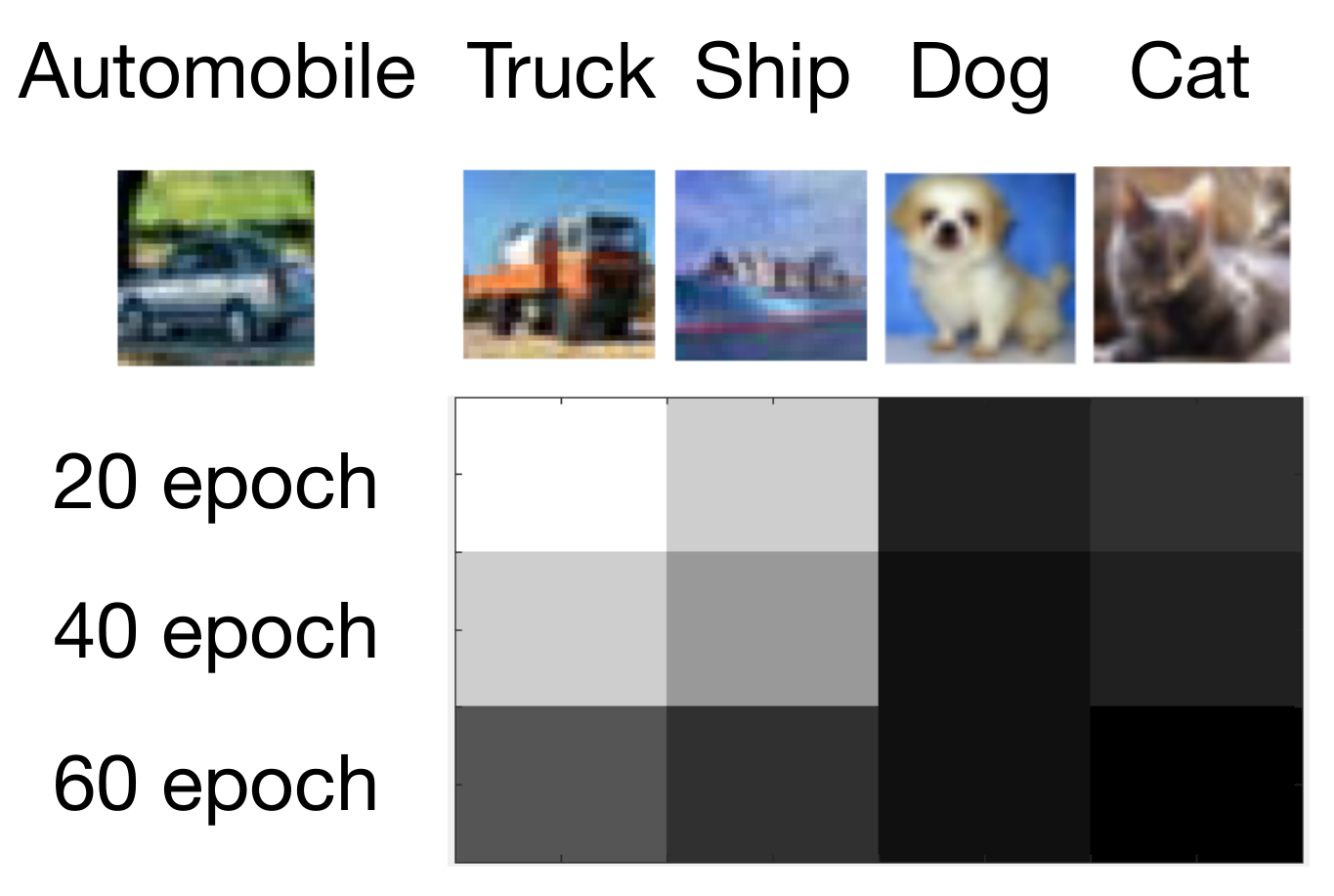}}
\caption{Evolution of class correlation scores $\Phi_t(i,j)$ on CIFAR-10 (with ResNet-32 network). Light/dark color denotes positive/negative values. Our policy modifies the class correlation scores in a way that forms a hierarchical classification curriculum by merging similar classes and gradually separating them.}
\label{fig:Cls_policy_visual}
\end{center}
\vskip -0.2in
\end{figure}

\textbf{Policy visualization for classification:} Figure~\ref{fig:Cls_policy_visual} illustrates the ALA policy learned for classification, which performs actions to adjust the loss parameters in $\Phi_t$ (\ie,~class correlations) dynamically. We observe that the ALA controller tends to first merge similar classes with positive $\Phi_t(i,j)$, and then gradually discriminates between them with negative $\Phi_t(i,j)$. This indicates a learned curriculum that guides model learning to achieve both better optimization and generalization.

\begin{figure}[!t]
\begin{center}
\centerline{\includegraphics[width=\columnwidth]{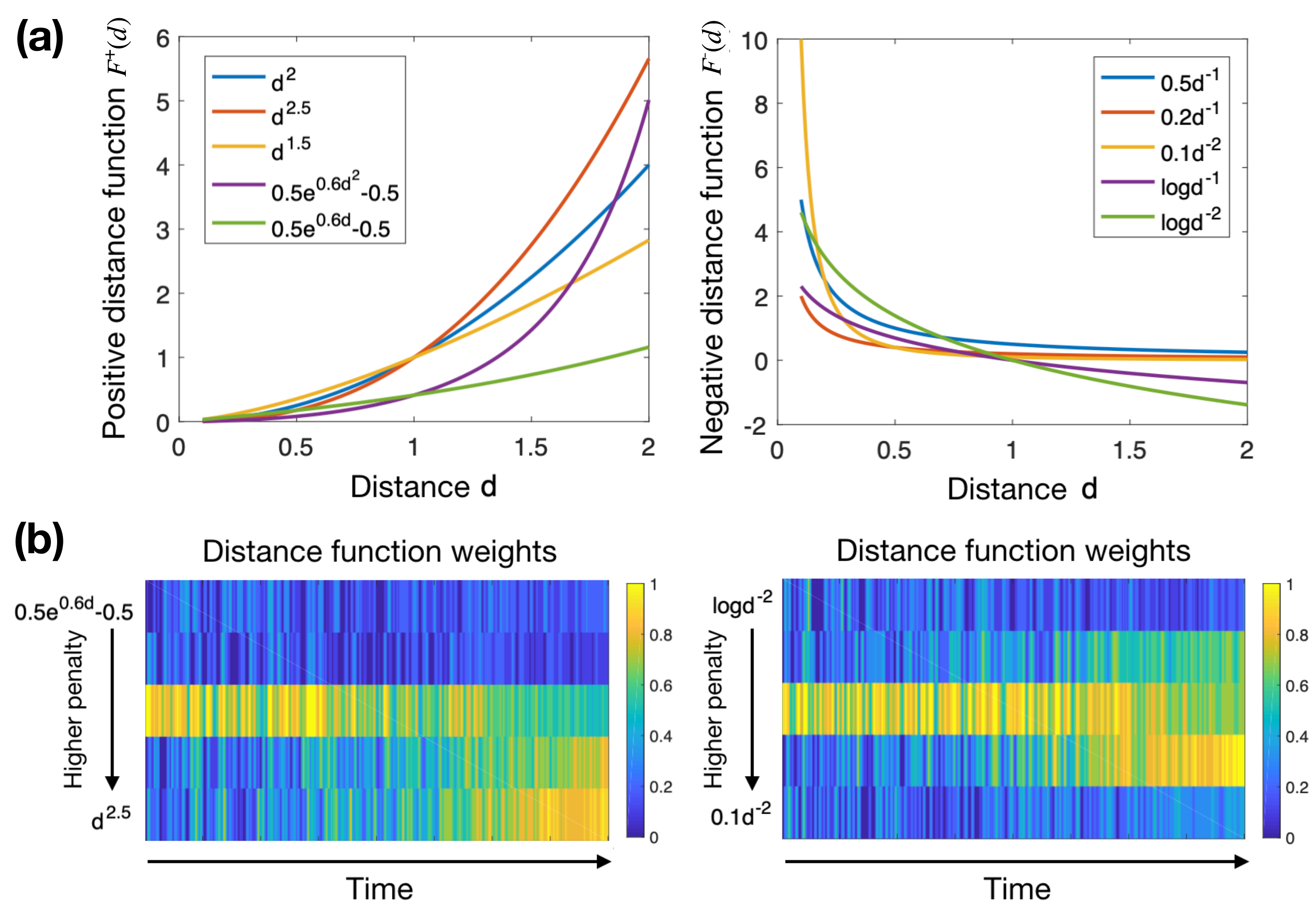}}
\caption{(a) Our positive distance function $F^+_i(\cdot)$ and negative distance function $F^-_i(\cdot)$ in metric learning. (b) Evolution of distance function weights $\Phi_t(i)$ on Stanford Online Products dataset. Our policy gives gradually larger weights to those high-penalty distance functions, which implies an adaptive and soft ``hard-mining'' curriculum.}
\label{fig:ML_policy_visual}
\end{center}
\vskip -0.3in
\end{figure}

\textbf{Policy visualization for metric learning:} We visualize the learned ALA policy for metric learning under a parametric loss formulation that mixes different distance functions. Figure~\ref{fig:ML_policy_visual}(a) first shows the distance functions $F^+_i(\cdot)$ and $F^-_i(\cdot)$ we apply to distance $d^+$ (between anchor and positive instances) and distance $d^-$ (between anchor and negative instances), respectively. Specifically, $F_i^+(d) \in \{ d^2, d^{2.5}, d^{1.5}, 0.5e^{0.6d^2} \!\!- 0.5, 0.5e^{0.6d} \!-\! 0.5\}$ defines 5 increasing distance functions to penalize large $d^+$, and~$F_i^-(d) \in \{ 0.5d^{-1}, 0.2d^{-1}, 0.1d^{-2}, \log d^{-1}, \log d^{-2}\}$ defines 5 decreasing distance functions to penalize small $d^-$. We empirically found our performance is relatively robust to the design choices of distance functions (within $\pm0.05\%$ verfication accuracy on LFW among our early trials), as long as they differ. The ability to learn adaptive weightings over these distance functions plays a more important role.

Figure~\ref{fig:ML_policy_visual}(b) demonstrates the evolution of weights $\Phi_t(i)$ over our distance functions on the Stanford Online Products dataset. Note that while the weights for our default distance functions $d^2$ and $0.5d^{-1}$ are both initialized as 1, our ALA controller learns to assign larger weights to those high-penalty distance functions over time. This implies an adaptive "hard mining" curriculum learned from data that is more flexible than hand-designed alternatives.

\section{Limitations}
In this work we studied multiple evaluation metric formulations (classification accuracy and AUCPR for the classification settings, and Recall@k and verification accuracy for metric learning). While this includes non-decomposable metrics, we did not extend to more complex scenarios that might reveal further benefits of ALA. In future work we plan to apply ALA to multiple simultaneous objectives, where the controller will need to weigh between these objectives dynamically. We would also like to examine cases where the output of a given model is an input into a more complex pipeline, which is common in production systems (\eg, detection$\rightarrow$alignment$\rightarrow$recognition pipelines). This requires further machinery to be developed for making reward evaluation efficient enough to learn the policy jointly with training the different modules.

Another area where ALA can be further developed is to make it less dependent on specific task types and loss/metric formulations. Ideally, a controller can be trained through continual learning to handle different scenarios flexibly. This would enable the use of ALA in distributed crowd learning settings where model training gets better and better over time.

Finally, an interesting area to study further is how ALA behaves in dynamically changing environments where available training data can change over time (\eg,  life-long learning, online learning, meta-learning). Ideally, ALA is suited to tackle these challenges, and we will continue to explore this in future work.

\end{document}